\pdfoutput=1

\documentclass[11pt]{article}

\usepackage[final]{acl}

\usepackage{times}
\usepackage{latexsym}
\usepackage{amsmath}
\usepackage{graphicx}
\usepackage{multirow}
\usepackage{graphicx}
\usepackage{hyperref}
\usepackage[T1]{fontenc}

\usepackage[utf8]{inputenc}

\usepackage{microtype}

\usepackage{inconsolata}

\title{Locating and Extracting Relational Concepts in Large Language Models}

\author{Zijian Wang \\
  University of Sydney \\
  \texttt{zwan0998@uni.sydney.edu.au} \\\And
  Britney Whyte \\
  BrewAI \\
  \texttt{britney@brewai.com} \\\And
  Chang Xu \\
  University of Sydney \\
  \texttt{c.xu@sydney.edu.au} 
  }

\begin{document}
\maketitle
\begin{abstract}
Relational concepts are indeed foundational to the structure of knowledge representation, as they facilitate the association between various entity concepts, allowing us to express and comprehend complex world knowledge.
By expressing relational concepts in natural language prompts, people can effortlessly interact with large language models (LLMs) and recall desired factual knowledge.
However, the process of knowledge recall lacks interpretability, and representations of relational concepts within LLMs remain unknown to us.
In this paper, we identify hidden states that can express entity and relational concepts through causal mediation analysis in fact recall processes.
Our finding reveals that at the last token position of the input prompt, there are hidden states that solely express the causal effects of relational concepts.
Based on this finding, we assume that these hidden states can be treated as relational representations and we can successfully extract them from LLMs.
The experimental results demonstrate high credibility of the relational representations: they can be flexibly transplanted into other fact recall processes, and can also be used as robust entity connectors.
Moreover, we also show that the relational representations exhibit significant potential for controllable fact recall through relation rewriting.
Check out the project on \url{https://github.com/Zijian007/Locate_Extract_Relation}.
\end{abstract}

\section{Introduction}
Pretrained GPT-based large language models (LLMs) can be regarded as knowledge bases, storing a large amount of knowledge about the real world. 
Each factual knowledge can usually be represented in the form of a triple $k =(s, r, o)$, where $s$ represents the subject entity, $o$ represents the object entity, and $r$ represents the relational concept that connects the subject and the object \cite{bollacker2008freebase,wang2017knowledge}.
For example, the fact that \textit{The capital of France is Paris} can be expressed as the triple \textit{(France, the capital of, Paris)}.
LLMs provide a flexible way to recall internal knowledge: people can use natural language prompts containing  subjects and relations to prompt LLMs, thereby stimulating the model responses containing corresponding objects 
\textit{i.e.}$(France, \hspace{3pt} the\hspace{3pt} capital\hspace{3pt} of) \to Paris$. \cite{geva2023dissecting,mann2020language}.
In this process, relationships play an important role: they connect paired entities to express various factual knowledge. 
Being able to locate and extract relational representations is of great value in exploring the interpretability of knowledge recall mechanism within LLMs.

\begin{figure}[t]
    \centering
    \includegraphics[scale=0.38]{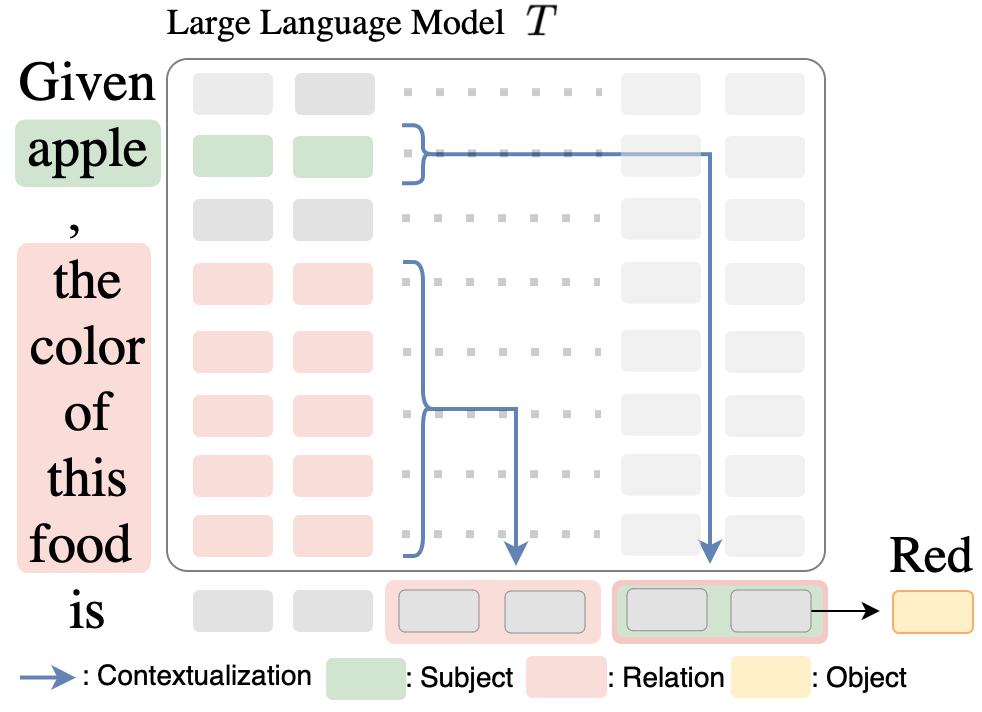}
    \caption{Our motivating observation in a fact recall process. At the last position, only hidden states in shallow layers solely express the relational causal effect, which provides us inspiration to treat these hidden states as relational representations.
    }
    \vspace{-2mm}
    \label{figure1}
\end{figure}

Recent studies on LLMs interpretability suggest that it is feasible to identify human-interpretable conceptual representations, including locating knowledge neurons \cite{meng2022locating,meng2022mass,dai2021knowledge,geva2020transformer,geva2022transformer}, finding semantic directions in activation space \cite{li2023inference,turner2023activation,zou2023representation}, 
extracting task vectors \cite{hendel2023context,todd2023function}, and approximating concept mapping \cite{hernandez2023linearity,hernandez2023inspecting}.
However, how relational concepts are represented within LLMs remains unexplored.
In this paper, we show that it is possible to locate and extract relational representations from LLMs, and these extracted representations can be connected with other entity concepts, enabling the expression of valuable knowledge.
Our research first reports an enlightening observation that provides us inspiration for locating relational representations during the course of casual mediation analysis\cite{vig2020investigating,meng2022locating}.
We find that at the last token position, the shallow hidden states solely present the causal effects of relations, and then deeper hidden states simultaneously express the causal effects of both subjects and relations in a more integrated manner, as shown in Figure \ref{figure1}.
Based on this finding, we hypothesize that at the last position, hidden states that only present relational causal effects can be regarded as Relational Representations and they can be extracted and reconnected with various other entities.
The reason why we focus on the last token is intuitive: due to the casual attention mechanism in GPT-based LLMs, only the last token captures complete contextual information to predict the object in the next token prediction way.


To verify this hypothesis, we conduct various experiments, including hidden states transplantation and zero-shot relational reasoning on data of multiple relation types.
The purpose of hidden states transplantation experiments is to verify the location accuracy of relational representations which includes a sliding pointer to specify the layer range for transplantation.
The results reveal that when the transplanted layer range only contains hidden states with relational causal effects, they exhibit a remarkable ability to express relational concepts, without absorbing any subject concepts.  
At this point, these states can be seamlessly transplanted to other fact-recall processes, enabling the recall of other relevant objects.
The purpose of zero-shot relational reasoning experiments is to verify the faithfulness and robustness of the relational representations that have been accurately extracted.
Our research demonstrates that the extracted relational representation acts as an effective entity connector, which is capable of accurately reasoning corresponding objects by taking various subjects as input.

We also expand the scope of our research into the more general fields of dialogue and text generation, where we try to use the relational representations to control LLMs responses.
Specifically, we propose the relation rewriting method, which can integrate other relational representations into the chatting process so that the model is required to generate content that aligns with the desired relation.

The main contributions can be summarized as follows:

$\bullet$ We observe that there are hidden states inside LLMs that solely express the causal effect of relations, which inspire us for locating the relational representations.

$\bullet$ We study the feasibility of extracting relational representations and verify that the extracted relation representations are very faithful and robust.

$\bullet$ We explore the potential of leveraging the relational representations for controllable LLMs response.

\section{Method}
In this section, we present the procedural details of how we derive our motivating observation and hypothesis for identifying relational representations through causal mediation analysis.

\subsection{Causal Mediation Analysis}
We first conduct a comprehensive examination of the hidden states that can express subject concepts and relational concepts during a successful fact recall process. 
This examination is carried out through causal mediation analysis, which is used by previous work to locate model parameters that can implicitly store knowledge \cite{meng2022locating,todd2023function}.
In this paper, we use this technique to gain a deep understanding of the distinctive roles played by hidden states in conveying subject and relational concepts in fact recall processes.

Specifically, causal mediation analysis models the information flow within LLMs as a causal graph, treating each hidden state as a causal mediating node. 
By measuring the causal effect of each node to a successful fact recall through a destruction-then-recovery paradigm, hidden states that play causal mediator roles can be located.


\begin{figure}[t]
    \centering
    \includegraphics[scale=0.37]{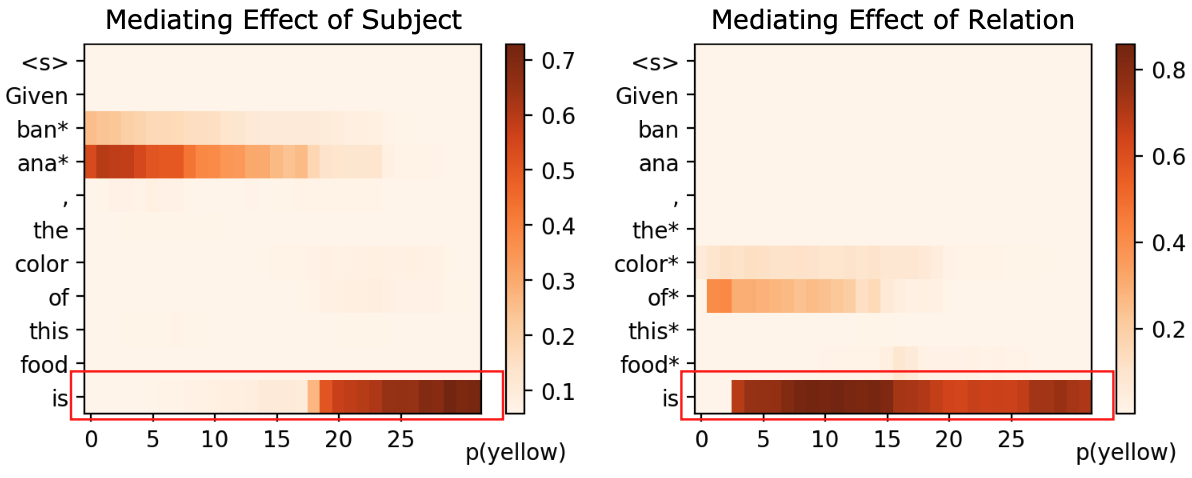}
    \caption{The Mediating Effect Visualization. We take \textit{"Given banana, the color of this fruit is"} as an example for illustration. We observe that at the last position, the mediating effect of the relation solely emerges in shallow layers, and then the mediating effect of the subject emerges in deep layers.}
    \vspace{-0mm}
    \label{ME_example}
\end{figure}

\begin{figure}[t]
    \centering
    \includegraphics[scale=0.27]{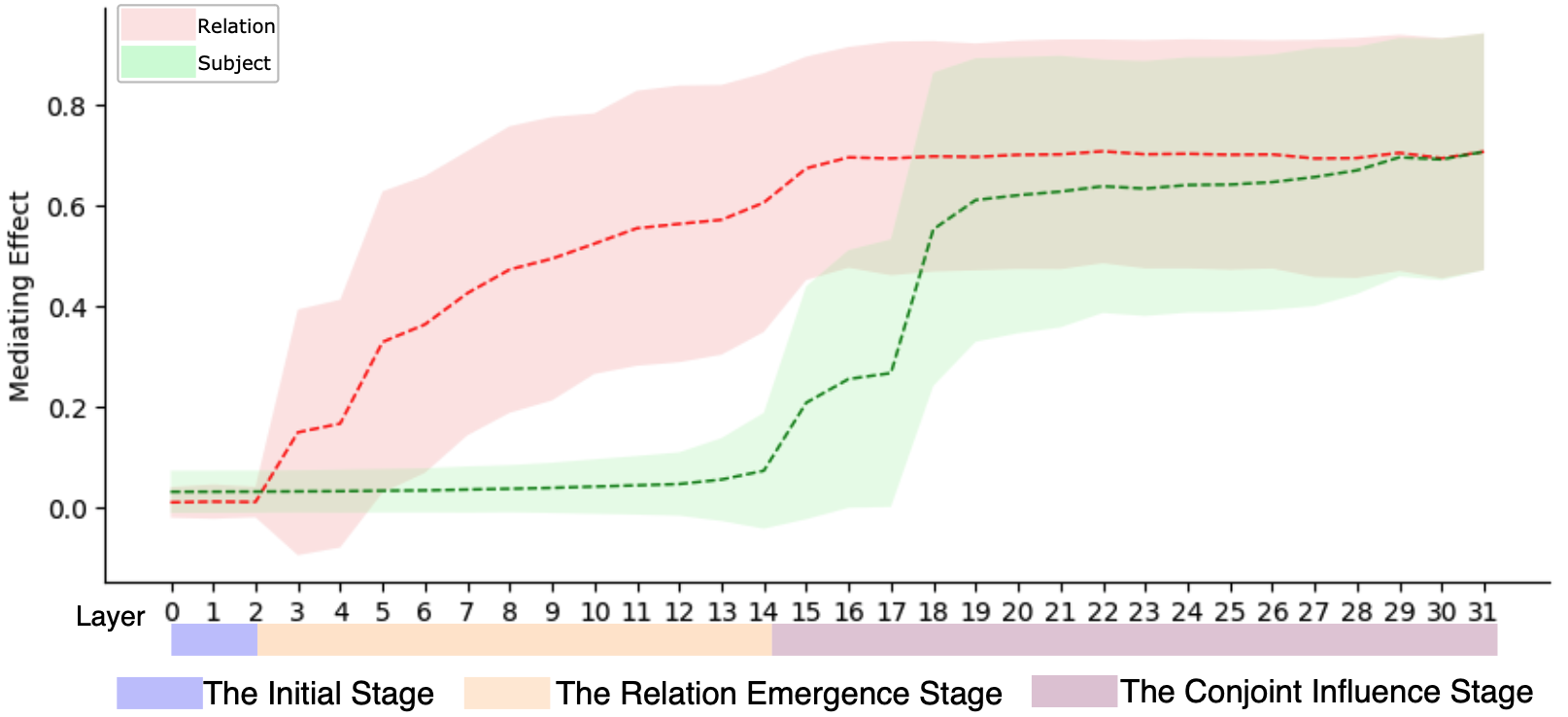}
    \caption{The average and variance area plot of the mediating effects of relations and subjects at the last position, across all relation types. We divide all layers into three stages to describe different patterns of causal effects of subjects and relations}
    \vspace{-3mm}
    \label{ME_plot}
\end{figure}

\subsubsection{Formulation}

An autoregressive transformer language model $T$ takes an input prompt $I(s,r)$ containing a subject entity and a relation concept, and outputs a probability distribution over its vocabulary set. 
Internally, the model contains $L$ layers to process the input sequence, including extracting subject and relation concepts and aggregating them to the last token for the next token prediction.
During the forward process, each hidden state $v^i_j$ serves as a casual mediator, where $i \in [1,H]$ denotes the token position, and $j\in[1,L]$ denotes the layer position.

When performing a fact recall process, the input $I(s,r)$ is first embedded into a sequence of vectors $v_0 = [v^1_0, v^2_0,...,v^H_0]$, which is processed by the subsequent multi-layer network.
Each layer includes a self-attention module $Self\_Att$ and a multi-layer perceptron $Mlp$ for hidden states processing. 
After multi-layer processing, hidden states at the last position $v^H_L$ receives subjects and relations information and are sent to a classification head $\phi$, followed by a softmax function, to calculate a distribution over the vocabulary. The object probability can be obtained when given the object id.
\vspace{-3mm}
\begin{align*}
     P_{\mathit{o}}(s,r)& = \mathit{Softmax}(\phi(v^H_L))[object\_id] \\
     v^H_j &= v^H_{j-1} +\\ 
     &\mathit{Self\_Att}(v^1_{j-1},v^2_{j-1},...,v^{H}_{j-1})+\\ 
     &\mathit{Mlp}(\mathit{Att}^H_{j}, v^{H}_{j-1}) \\
\end{align*}
\vspace{-6mm}
\subsubsection{Analysis Method}

Taking subject analysis as an example, we adopt a paradigm of destruction first and then recovery in our analysis \cite{meng2022locating,vig2020investigating}.
Specifically, we first collect and record all hidden states $v^i_j$ in a successful object prediction, which will be used for state restoring in the recovery step. 
Then in the destruction step, we introduce noise to embedding vectors of subject tokens, which contaminate all subsequent hidden states in the model.
This contamination disrupts the information flow and makes the model's prediction probability for the correct object $P_{\mathit{o}}(s^*,r)$ has a notable decrease, where $s^*$ means the subject is contaminated.

In the subsequent recovery step, we restore each contaminated hidden state node back to its original value one by one, using the clean states previously recorded.
It is crucial to note that throughout the recovery process, the subject embedding vectors remain contaminated, maintaining the disruption introduced in the destruction step. 
Following restoring each hidden state, we record the model's predicted probability for the correct object $P^{restore}_{\mathit{o}}(s^*,r)$.

We define the difference value observed in the predicted probability for the correct object as the Mediating Effect of the corresponding hidden state node \textit{i.e.}$\mathrm{ME}(s) = P^{restore}_{\mathit{o}}(s^*,r)-P_{\mathit{o}}(s^*,r)$.
By measuring the mediating effect, we can identify which hidden state successfully encodes the subject concept thereby contributing to the accurate prediction of the object.

In the analysis of relations, a similar pipeline is applied, except that the destroyed part becomes the relation token embedding.
Then the corresponding mediating effect of relation $\mathrm{ME}(r) = P^{restore}_{\mathit{o}}(s,r^*)-P_{\mathit{o}}(s,r^*)$ is calculated in the recovery step.
\vspace{-1mm}

\subsubsection{Data}
\label{2.1.3}
We construct a database with 3500 subject-object pairs, spanning 22 relation types, filtered from \cite{hernandez2023linearity}.
We use the following prompt template: 
\textit{"Given $<$subject$>$, the $<$relation$>$ of this one is"}, which is proved successful in prompting the model to predict the corresponding object in the next token prediction manner.
More dataset construction details and template options are discussed in the Appendix.

\subsection{A Three-Stage Observation}

Figure \ref{ME_example} provides a visualized result of causal mediation analysis.
For illustrative purposes, we take the prompt \textit{"Given banana, the color of this food is"} as an example.

It is not a novel discovery that hidden states at both the subject and relation tokens' positions express obvious effects.
This phenomenon, commonly known as concept enrichment, aims to encode more relevant attributes to facilitate subsequent processing \cite{meng2022locating,geva2023dissecting}.

However, when we have a more detailed examination at the last position of the whole prompt sequence, we observe distinct patterns in the mediating effects of subjects and relations. 
As shown in Figure \ref{ME_plot}, we plot the average mediating effect of subjects and relations across all relations at the last position, using Mistral-7b-instruct model \cite{jiang2023mistral}.
All layers can be categorized into three stages to describe these patterns: the Initial stage, the Relational Emergence stage, and the Conjoint Influence stage.

$\bullet$ In the Initial stage, typically spanning 0-3 layers, the hidden states do not show any mediating effects of relations and subjects.

$\bullet$ In the Relational Emergence stage, typically spanning 4-15 layers, hidden states only show the mediating effects of relations rather than the effects of subjects.

$\bullet$ In the Conjoint Influence stage, typically spanning 16-31 layers, hidden states express both relational and subject effects, indicating their common influence on object prediction.

\subsection{Our Hypothesis for Relation Locating}
Based on the above discoveries, we propose a hypothesis that \emph{hidden states located in the Relational Emergence stage can be considered as relational representations}, because these hidden states exclusively manifest relational causal effects, showcasing their specialized role in encoding and expressing the relational concept.

\section{Experiments For Hypothesis Validation}
We design experiments from two aspects to verify our hypothesis.
The hidden states transplantation experiment aims to prove that the hidden states in the Relational Emergence stage indeed accurately locate the relational representation, and the zero-shot relational reasoning experiment aims to study the faithfulness and robustness of the relational representations.

\subsection{Hidden States Transplantation}
The idea behind this experiment is intuitive: if hidden states in the Conjoint Influence stage are also considered as the relational representations, then the subject information will also be included and extracted together, which could lead to the entanglement of the relational representations and the subject information.

Formally, given a reference prompt $I(s_1,r_1)$ with an object $o_1$, and a source prompt $I(s_2,r_2)$ with an object $o_2$, 
we aim to transplant the relational representations from $I(s_1,r_1)$ to $I(s_2,r_2)$.
Ideally, the model should be able to predict a target object $o_3$ corresponding to the new subject-relation pairs \textit{i.e.}$(s_2,r_1) = o_3$.
However, if the reference subject information $s_1$ is also transplanted,
the prediction result is switched to $o_1$ \textit{i.e.}$(s_1,r_1) = o_1$.

\subsubsection{Implementation}
\begin{figure}[t]
    \centering
    \includegraphics[scale=0.25]{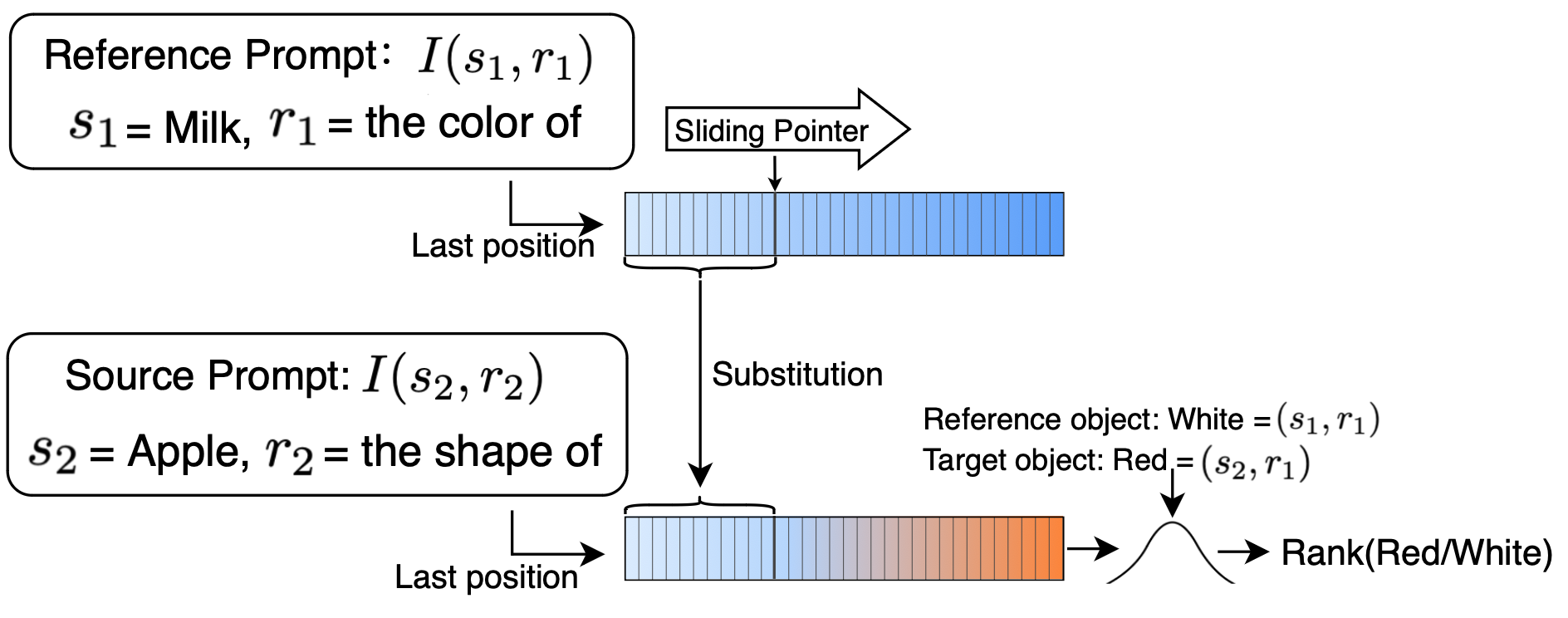}
    \caption{The illustration of hidden States transplantation, which includes a sliding pointer to dynamically indicate the layer range.}
    \vspace{-2mm}
    \label{transplant_figure}
\end{figure}

\paragraph{Transplantation details}
The illustration of the experimental process is shown in Figure \ref{transplant_figure},
To dynamically adjust the layer range of transplantation, we design a method that involves sliding an end pointer at the last position to indicate the layer range. 
The sliding of the end pointer starts from the first layer and gradually moves towards the last layer. 
Every time the pointer slides, we transplant all hidden states preceding the pointer position from the reference to the source process.
Following each transplantation, we record the predicted rank of the target object $o_3$ as well as the reference object $o_1$.
By analyzing trends in ranking changes, we can ascertain whether the transplanted hidden states solely encode the relational concept or also encode information about the subject. 

\begin{figure*}[t]
    \centering
    \includegraphics[scale=0.37]{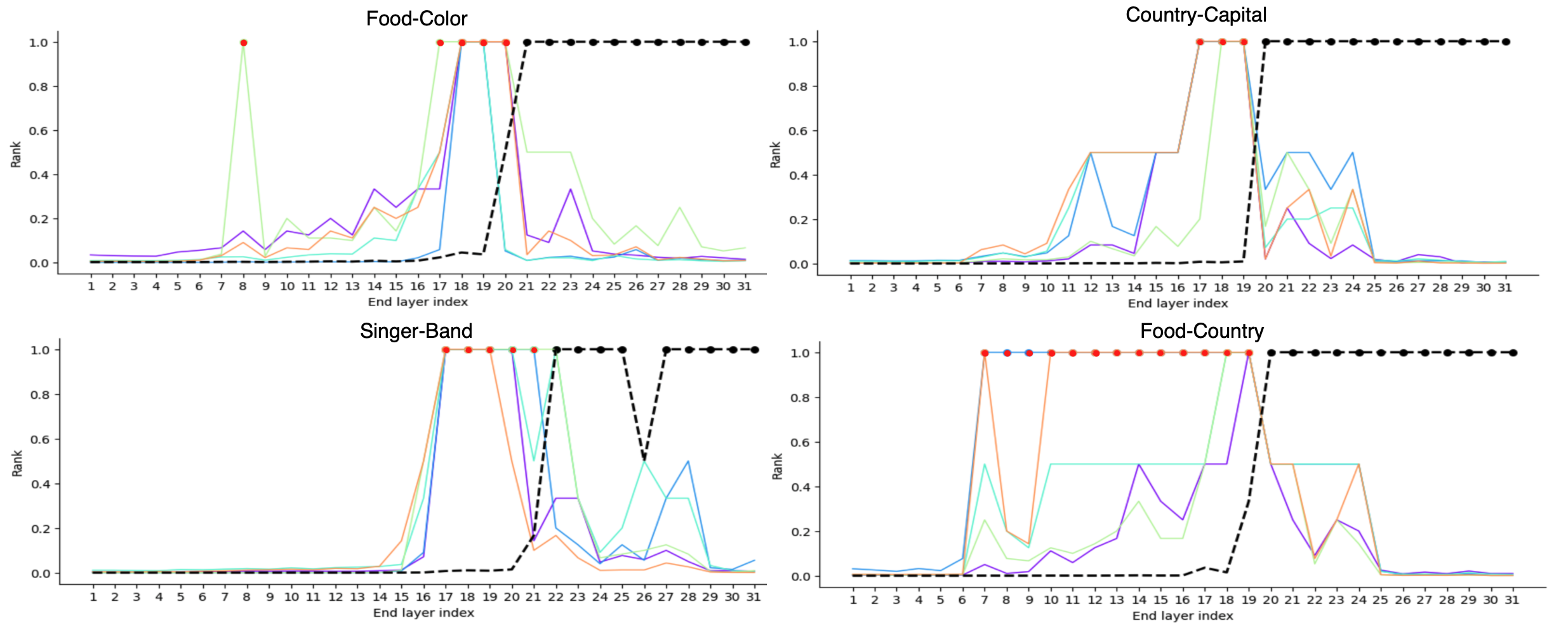}
    \caption{The prediction rank reciprocal of target and reference objects. We select 4 relation types for illustration. Red circles denote successful predictions of target objects, while black circles denote successful predictions of reference objects. The colorful lines represent 5 target object predictions, and the black line represents the reference object prediction.}
    \vspace{-3mm}
    \label{rank}
\end{figure*}

\paragraph{Prompts construction}
For our analysis, we carefully curate data from 10 relation types, with each type comprising multiple reference-source pairs.
When constructing reference prompts, we select one subject-relation pair from the dataset and fill them into the template. 
When constructing source prompts, we combine the subject with other relations. 
The choice of source relation is not important since it will be covered by the reference relation in the experiment.

There are two things to note when building these prompts. 
First, the reference subject and source subjects should have no intersection, so as to avoid being unable to determine whether subject information is leaked.
Second, the model should be able to complete object predictions for reference and source prompts to ensure that relevant knowledge has been stored in the model. 
More details are discussed in the Appendix.

\subsubsection{Result and Analysis}
In Figure \ref{rank}, we select 4 relation types as examples to interpret the experimental results. 
This figure shows the model’s prediction of the target object and reference object following each hidden state transplantation.
The x-axis coordinate represents the end point position and the y-axis coordinate represents the \emph{reciprocal of rank}. 
If the \emph{reciprocal of rank} is equal to 1, it means that the object has been successfully predicted.
The colored lines and black lines represent the model's predictions of the target object $o_3$ and the reference object $o_1$ after each relationship transplantation, respectively.
We utilize red circles to denote successful predictions of the target object, while black circles indicate successful predictions of the reference object. 
Table \ref{table_acc_transplant} also shows the average accuracy of the prediction of the target and the reference object when the end pointer is located in different layer ranges.

\paragraph{Validity of the Three-Stage Observation}
It is evident that as the endpoint layer moves, there are corresponding changes in the model's predicted ranking and accuracy of the target object and reference object.
When the endpoint layer is situated in the Initial stage, the model does not provide any predictions related to the target or reference objects. 
Table \ref{table_acc_transplant} also shows that in the initial stage, neither target objects nor reference objects can be predicted.
This is because, in the Initial stage, hidden states do not encode any concepts of relations and subjects.

In the case where the ending layer is located in the Relational Emergence stage, particularly towards the end of this stage, the model demonstrates a remarkable ability to predict the target object $o_3$, as shown in the red circles. 
The prediction accuracy of reference objects has also increased significantly in Table \ref{table_acc_transplant}.
This observation implies that the relational concept of the reference prompt is effectively transplanted to the source prompt, while the subject concept of the reference prompt shows no discernible impact. 

When the end pointer enters the Conjoint Influence stage, the model suddenly fails to predict the target object $o_3$, while it starts to successfully predict the reference object $o_1$, as shown in the black circle in the Figure 5.
This observation can be attributed to the joint transplantation of the information from both the reference subject $s_1$ and the reference relationship $r_1$, which leads to the prediction of $o_1$.

Based on the comprehensive analysis, we believe that hidden states within the Relation Emergence stage effectively encode the relationship concept without any leakage of subject information.

\begin{table}[]
\caption{The prediction accuracy of target/reference objects when the end pointer is located in different layer ranges.}
\label{table_acc_transplant}
\resizebox{0.5\textwidth}{!}{
\begin{tabular}{ccccc}
\hline
\multicolumn{1}{c|}{\multirow{2}{*}{\small Subject-Object}} & \multicolumn{4}{c}{\small Layer range}         \\ \cline{2-5} 
\multicolumn{1}{c|}{}         & {\small 1-4} & {\small 5-8}  & {\small 9-17} & {\small 18-32}         \\ \hline
{\small Country-Capital}       & {\small 0/0 }& {\small 7.5\%/0} & {\small 88.3\%/3.5\% }& {\small 3.3\%/94.1\%} \\
{\small Food-Color}        &{\small 0/0 }& {\small 5\%/0 }  & {\small 84.7\%/5.9\% }& {\small 2.4\%/93.3\%} \\
{\small County-language  }      & {\small 0/0 }& {\small 4.8\%/0 }& {\small 85.5\%/4.1\% }& {\small 2.6\%/97.5\%} \\
{\small Athlete-Sport   }            & {\small 0/0 }&{\small 5.6\%/0}& {\small 82.3\%/2.9\% }& {\small 5.3\%/88.3\%} \\
{\small Person-Religion  }        & {\small 0/0} & {\small 4.4\%/0 }& {\small 88.3\%/5.1\% }& {\small 3.5\%/87.3\% }\\
{\small Product-Company  }        & {\small 0/0} & {\small 8.3\%/0} & {\small 82.3\%/8.5\% }& {\small 2.9\%/92.9\% }\\
{\small Singer-Band     }        &{\small 0/0 }& {\small 6.5\%/0 }& {\small 86.7\%/4.5\% }& {\small 4.3\%/93.3\% }\\
{\small Food-country     }        & {\small 0/0} & {\small 7.4\%/0 }& {\small 85.6\%/5.5\% }& {\small 3.5\%/95.6\% }\\
{\small Person-Instrument}       & {\small 0/0 }& {\small 2.1\%/0} & {\small 87.5\%/4.8\% }& {\small 5.6\%/93.3\%} \\
{\small Country-Currency  }     & {\small 0/0 }& {\small 5.0\%/0 }& {\small 91.3\%/4.1\%} & {\small 6.3\%/95.3\% }\\ \hline
\end{tabular}
}
\end{table}

\begin{figure*}[]
    \centering
    \includegraphics[scale=0.39]{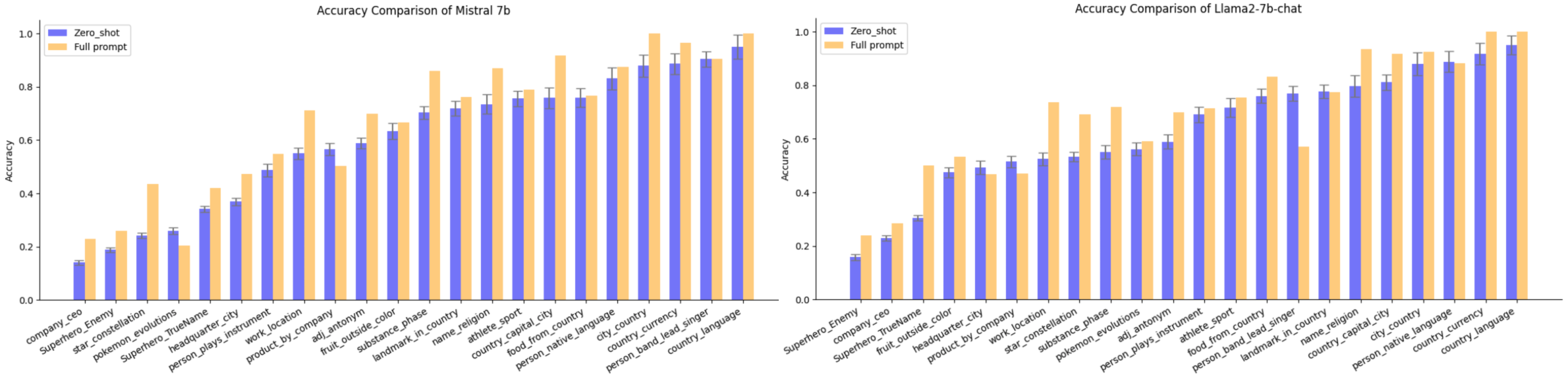}
    \caption{Average accuracy comparison across all subject entities in each relation type for Mistral-7b-instruct and Llama2-7b-chat model, under the zero-shot and full prompt settings, respectively.}
    \vspace{-0mm}
    \label{acc_compare}
\end{figure*}

\subsection{Zero-shot Relational Reasoning}
Once the relational representation is accurately located and extracted, it becomes imperative to study its faithfulness and robustness.
In this section, we regard the relational representations as entity connectors $E(r)$, which enable LLMs to reason the corresponding objects when the model takes prompts only containing subject entities as input.
By measuring the accuracy of the model's reasoning about the object, the faithfulness of the relational representation can be evaluated.

Previous work performs similar studies on task vectors within LLMs \cite{hendel2023context,todd2023function}, but these studies are conducted in the few-shot setting of in-context learning.
In contrast, our study was conducted under the zero-shot relational reasoning setting.


\subsubsection{Implementation}
To perform zero-shot relational reasoning, we allow the model to only take prompts that contain subject information as input.
The template for input prompts is \textit{"Given $<$subject$>$,"} or \textit{"$<$subject$>$$\rightarrow$"}.
During forward inference, we insert the entity connector $E(r)$ in the Relational Emergence stage at the last position, and examine the output, which is $\hat{o} = T(I(s), E(r))$.
Then the faithfulness can be evaluated by the matching accuracy between the reasoned objects $\hat{o}$ and ground truth objects $o$.


\subsubsection{Results and Analysis}
\paragraph{Relational Reasoning Accuracy}
The experimental results are shown in Figure \ref{acc_compare}. 
We conduct a comparison between the zero-shot accuracy and the accuracy obtained using the regular full prompt mentioned in Section \ref{2.1.3}, using the Mistral-7b-instruct and Llama2-7b-chat model\cite{touvron2023llama}.
We observe that, except in some complex relationship types, such as \textit{star constellation} and \textit{company ceo}, the zero-shot accuracy did not significantly decrease compared to the accuracy obtained using the regular full prompt. 
Moreover, in certain individual cases, we even observe that the zero-shot learning performs better than the regular full prompt.
These results strongly suggest that the extracted relational
representations exhibit a high level of faithfulness to serving as entity connectors.

\paragraph{Robustness}
Since the entity connector $E(r)$ is extracted from the prompt $I(s,r)$, it becomes crucial to assess the robustness of $E(r)$ to variations of the subject $s$. 
To verify this, we extract $E(r)$ from various $I(s,r)$ containing different subjects, and calculate the standard deviation of the relational reasoning accuracy.
As shown in the grey error area in Figure \ref{acc_compare}, the floating range of accuracy is relatively stable.

\begin{figure}[]
    \centering
    \includegraphics[scale=0.27]{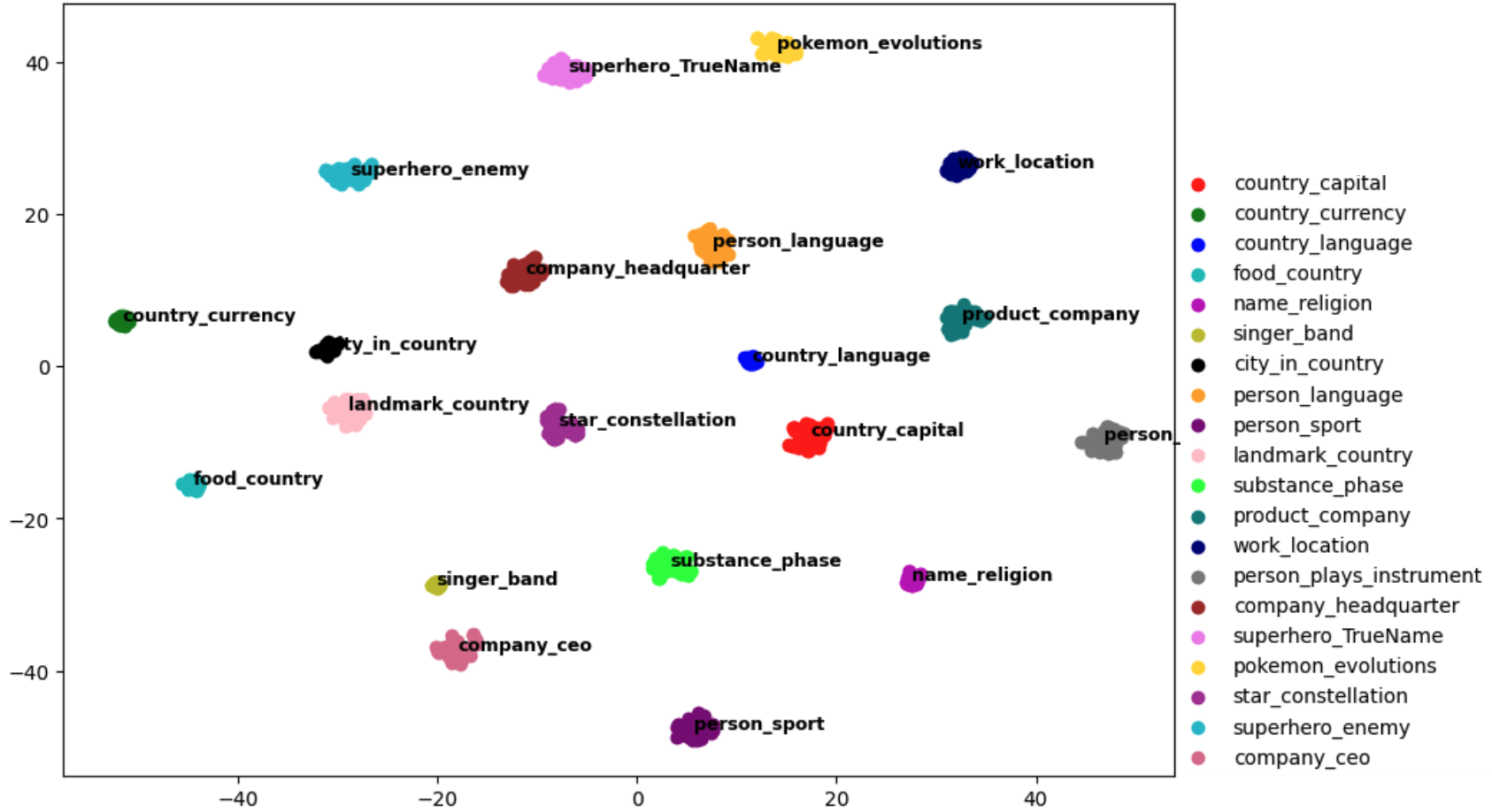}
    \caption{A t-SNE plot of various relational representations. Each color represents a relationship type and each data point is 
    derived from different subjects using Mistral 7b. We can observe that each relation forms its own distinct cluster.}
    \vspace{-0mm}
    \label{tsne}
\end{figure}

We also perform a visual analysis of the geometric structure of various $E(r)$ by using t-SNE dimensionality reduction, as shown in Figure \ref{tsne}.
We find that distinct clusters formed between different types of attribute extractors, with significant inter-class distances and small intra-class distances.

\section{Application in Controllable Fact Recall}
After the above discussion, we have proved that the relational representation can be extracted from LLMs, with great faithfulness and robustness.
In this section, we aim to utilize this property to achieve controllable fact recall by taking advantage of relational representation in more general scenarios.

\subsection{Method}
\begin{figure}[]
    \centering
    \includegraphics[scale=0.32]{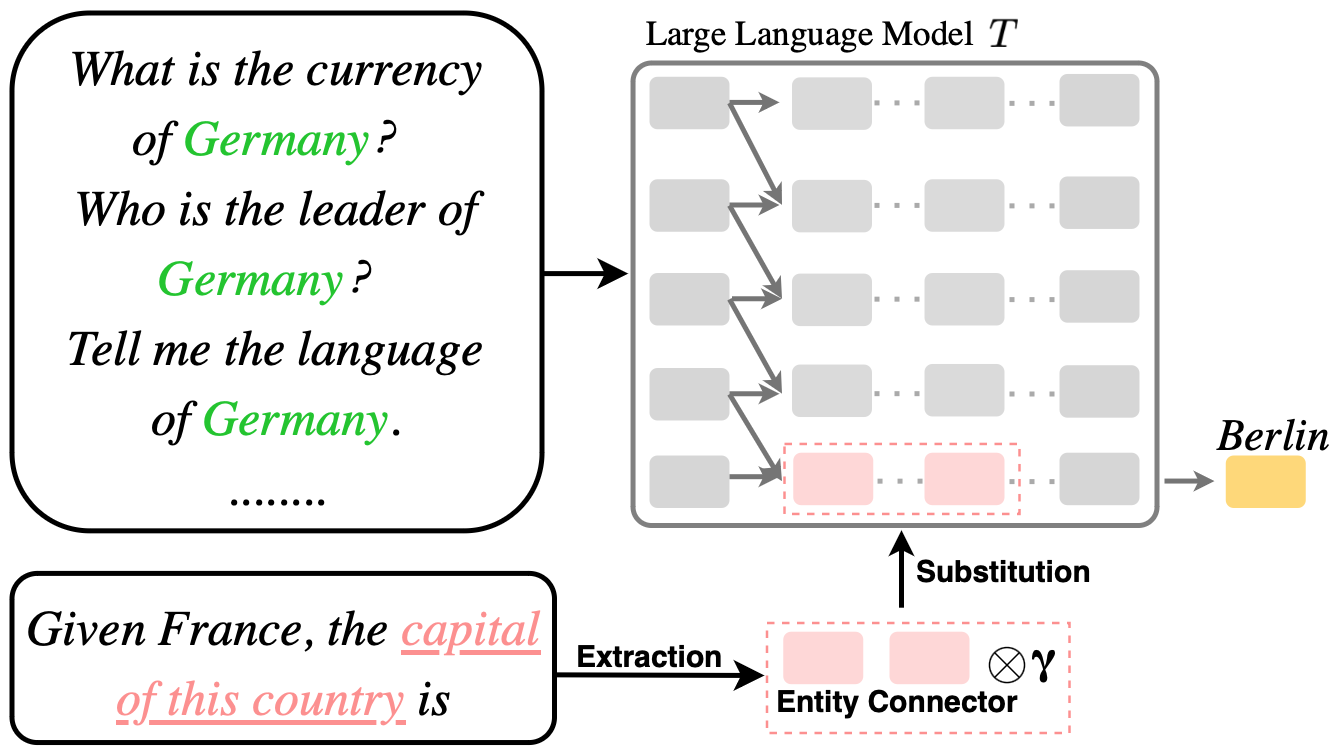}
    \caption{The illustration of our relation rewriting method, which can rewrite relational concepts contained in the input inquiries.}
    \vspace{-0mm}
    \label{control_figure}
\end{figure}

The idea of our method is very intuitive: when a user tries to recall factual knowledge by chatting with LLMs, we can rewrite the original relation concept inside the model to elicit other object predictions.
As illustrated in Figure \ref{control_figure}, when we pose queries to LLMs, such as \textit{"What is the currency of Germany?"} or other related factual inquiries about the same subject, we have the flexibility to insert another relational representation.

For example, when we insert the relation concept \textit{"the capital of"}, a new connection \textit{(Germany, the capital of)} is established.
As a result, the model is guided to generate outputs that align with the inserted relation, leading to the desired output of \textit{"Berlin"}. 
This method can be referred to as \textbf{"relation rewriting"} as it comprehensively covers the relational concept present in the original prompt. 
Also, during insertion, a multiplicative factor $\gamma$ is used to control the rewriting strength.

We are also interested in whether we can add new relation descriptions directly to obtain a new input prompt $I(s,r',r^t)$ for relation rewriting. 
Specifically, we can add a description of the new relation after the original inquiry: \textit{"What is the currency of Germany? Actually, I am asking the capital city."}
We compare this method, which serves as a baseline, with our proposed relational rewriting method.

\subsection{Experiments}
We manually construct subject-relation pairs of 12 relation types to verify the relation rewriting method.
For each subject entity $s$, we randomly pair it with different relations $r'$ and fill them into various query templates to construct new inquiry prompts $I(s, r')$, which need to be rewritten relationally.
More data details are discussed in the Appendix.
It is worth noting that some subject-relation combinations may be semantically meaningless or lack a clear association, such as \textit{(Germany, the taste of)}, but our approach still needs to rewrite these relations.
Then we input these inquiry prompts to LLMs, and insert another relational representation in the inference process to check whether the outputs are match with the inserted relation.

\subsection{Results}
The relation rewriting result is shown in Table \ref{table2_rewrite}.
We show the average accuracy across all subjects within each relation and compare the accuracy of different $\gamma$ and the baseline method.
We can observe that the proposed relation rewriting method performs much better than directly modifying prompts in all relation types.
This suggests that modifying relational concept at the representation level is more efficient than modifying relational representations in the prompts.
Moreover, in some cases, accuracy improves with increasing the rewriting strength $\gamma$.

We also apply the proposed relation rewriting method to open-ended text generation scenarios to narrow the model’s response range to the desired relation.
The results are shown in the Appendix.

\begin{table}[h]
\caption{Relation rewriting accuracy comparison with the baseline method and different rewriting strength $\gamma$.}
\label{table2_rewrite}
\resizebox{0.5\textwidth}{!}{
\begin{tabular}{lllll}

\hline
                  & \multicolumn{1}{c}{$\gamma$=1} & \multicolumn{1}{c}{$\gamma$=1.5} & \multicolumn{1}{c}{$\gamma$=2} & Modify Prompt\\ \hline
Country-Capital   & 86.5\%  & 86.5\%    & 86.5\%   & \multicolumn{1}{c}{15.8\%} \\
Food-Color        & 80.6\%  & 80.6\%    & 80.1\%   & \multicolumn{1}{c}{10.1\%} \\
Country-Currency  & 89.8\%  & 89.8\%    & 89.8\%   & \multicolumn{1}{c}{12.6\%} \\
Country-Language  & 87.2\%  & 88.5\%    & 88.5\%   & \multicolumn{1}{c}{13.5\%} \\
Singer-Band       & 85.5\%  & 85.5\%    & 87.1\%   & \multicolumn{1}{c}{9.5\%} \\
Athlete-Sport  & 91.8\%  & 92.1\%    & 92.1\%   & \multicolumn{1}{c}{4.4\%} \\
Food-Country  & 90.3\%  & 90.3\%    & 90.9\%   & \multicolumn{1}{c}{1.8\%} \\
Superhero-True name & 71.5\%  & 72.3\%    & 72.3\%   & \multicolumn{1}{c}{11.9\%} \\ 
Superhero-Enemy & 72.5\%  & 72.5\%    & 73.1\%   & \multicolumn{1}{c}{8.3\%} \\ 
Company-CEO & 79.5\%  & 79.8\%    & 79.8\%   & \multicolumn{1}{c}{8.8\%} \\ 
Company-headquarter & 85.1\%  & 87.4\%    & 86.5\%   & \multicolumn{1}{c}{5.7\%} \\ 
\hline
\end{tabular}
}
\end{table}

\section{Limitations}
\begin{figure}[t]
    \centering
    \includegraphics[scale=0.35]{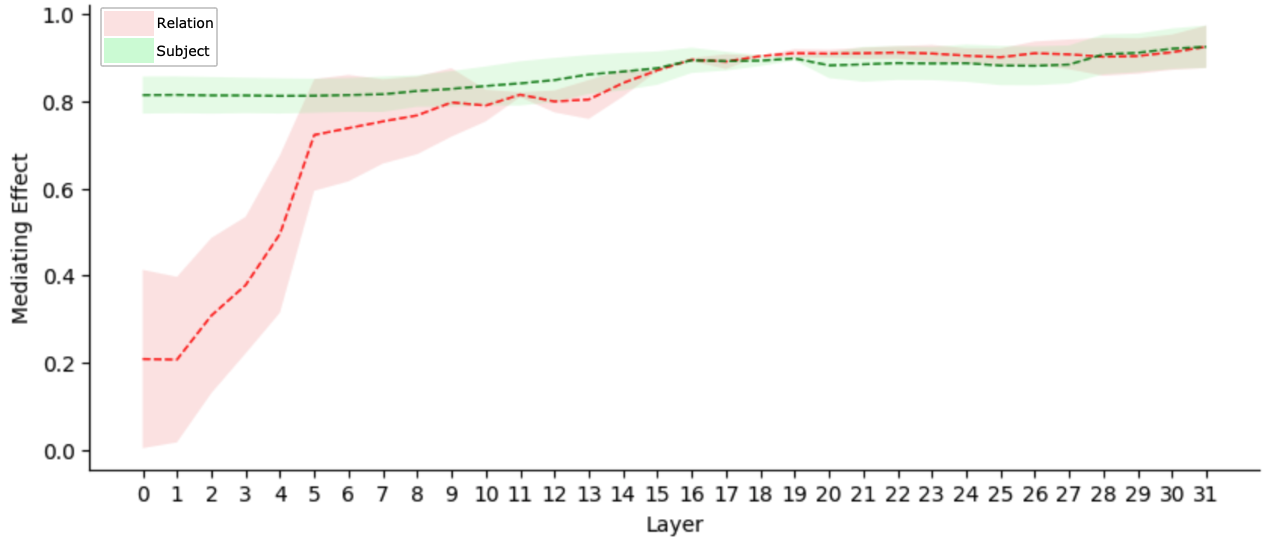}
    \caption{The average casual mediating effects of subjects and relations in linguistics-related tasks. We observe that the effects of subjects emerge first, which is inconsistent with our three stage obseravtion.}
    \vspace{-0mm}
    \label{limit_linguist}
\end{figure}

Although the three-stage observation provides us inspiration for locating relational representation,
this observation is uncommon in fact recall processes involving linguistics.
For example, when LLMs predict the past tense or comparative form of a word, the causal effects of subjects emerge before relations, which is inconsistent with our three-stage observation.
As shown in Figure \ref{limit_linguist}, we collect the average causal effects of subjects and relations at the last position in a number of linguistic tasks.
We find that the causal effect of subjects is very obvious from the first layer, while the causal effects of relations still have a gradual rising process.
We speculate that this is because linguistic tasks only entail the application of predetermined grammatical rules, without requiring multi-layer processing to capture the semantics of words.
This limits our method to linguistic relation location and extraction.

\section{Related work}
\subsection{Knowledge Probing in Language Models}
Probing the internal knowledge of neural networks is a fundamental problem.
In LLMs, numerous studies have been conducted by training a probing classifier in activation space, which could discover some meaningful concept directions
\cite{marks2023geometry,park2023linear,burns2022discovering,li2023inference,turner2023activation,zou2023representation,de2021editing}.
Compared to these, our method avoids the training process and we can easily extract and transplant the concept representations.

\subsection{Knowledge Editing in Language Models}
Many efficient knowledge editing methods have been proposed to ensure the integrity and real-time nature of LLMs
\cite{mitchell2022memory,madaan2022memory,huang2023transformer,de2021editing,mitchell2021fast,gupta2023editing,dai2023neural}. 
These methods usually require updating model parameters or introducing external parameters to ensure that LLMs generate optimal answers.
However, our method can guide the model to predict another answers just by replacing the relational representations during the inference process, without any parameter modification.

\section{Conclusion}
In this paper, we analyze the fact recall process within GPT-based LLMs.
Our investigation reveals that within LLMs, there are specific hidden states that can solely express relation concepts, and we hypothesise that these hidden states can be regarded as relational representations.
We design hidden states transplantation experiments and zero-shot relational reasoning experiments to verify our hypothesis.
Experimental results show that these hidden states indeed accurately locate relational representations without absorbing other entity concepts, and they can be extracted as entity connectors with high faithfulness and robustness.
Based on the above findings, we explore the application potential of the relational representations in controllable LLMs response.
We leverage the relational representations to intervene with user inquiry requests for LLMs, and successfully guide the model to output other answers.
Our work sheds light on the study of interpretability of LLMs and provides a stepping stone towards understanding the internal knowledge system of LLMs.


\section{Acknowledgement}
This work was fully funded by BrewAI, through a PhD program. We would like to express our sincere gratitude to BrewAI and its CEO, Gavin Whyte, for their invaluable support of this research project.
BrewAI is a leading company in the field of Generative Artificial Intelligence, dedicated to providing automated AI services to companies. Their expertise and commitment to research innovation provided a critical foundation for this research project.

\newpage

\bibliography{main}

\appendix

\section{Appendix}
\label{sec:appendix}

\subsection{Controllable Open-ended Text Generation Results}
In this section, we introduce the application of our proposed relation rewriting method in the Open-ended Text Generation scenario.
Specifically, when we prompt LLMs to generate some information about a subject, we can insert a relational representation in the inference process, to narrow the model's response range to the specified relation.
The results are shown in Table \ref{generation result1}, \ref{generation result2}, \ref{generation result3}, \ref{generation result4}.
We observe that before the relation is rewritten, LLMs introduce the subject information broadly, without a specific scope.
However, after the relation is rewritten, the information introduced by the LLMs is limited to a specific scope, indicating that relationship rewriting can guide the model's response range.

\subsection{The Three-Stage Observation in Llama.}
\begin{figure}[h]
    \centering
    \includegraphics[scale=0.27]{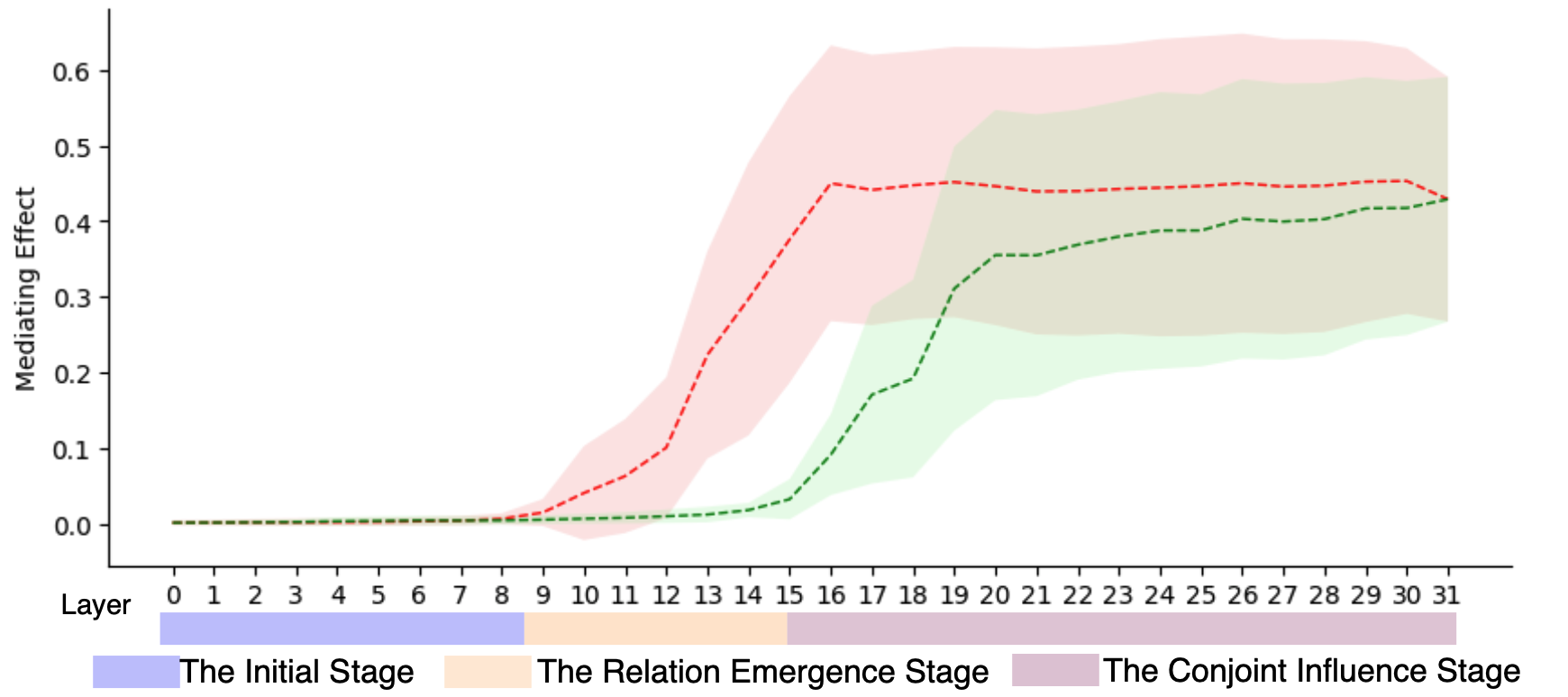}
    \caption{The last position mediating effects of subjects and relations of llama-7b-chat. The Three-Stage Observation is still valid.}
    \label{figure_me_llama}
    \vspace{-0mm}
\end{figure}

We discuss whether the Three-Stage Observation also exists in Llama-7b-chat model.
Figure \ref{figure_me_llama} shows the last position mediating effects of subjects and relations of llama-7b-chat model. 
All layers can also be divided into three stages to describe different causal effect patterns of subjects and relations.

$\bullet$ In the Initial stage, typically spanning 0-8 layers, the hidden states do not show any mediating effects of relations and subjects.

$\bullet$ In the Relational Emergence stage,typically spanning 9-15 layers, hidden states only show the mediating effects of relations, rather than the effects of subjects.

$\bullet$ In the Conjoint Influence stage, typically spanning 16-31 layers, hidden states express both relational and subject effects, indicating their common influence on object prediction.

\subsection{Influence about Various Prompt Templates}
In this study, we investigate the influence of various prompt templates on the causal effects of subjects and relations in the process of fact recall. 
Specifically, we focus on analyzing the impact of the order of subjects and relations within the prompts.
In our previous template \textit{"Given $<$subject$>$, the $<$relation$>$ of this one is"}, subject appears before relation.
In order to present the subject after the relation and enable object prediction in the next token prediction approach, we propose two new templates. 
The first one is \textit{"If I want to know the $<$relation$>$ of $<$subject$>$, the answer is"}, and another is \textit{"Tell me that the $<$relation$>$ of $<$subject$>$ is"}.

\begin{figure}[t]
    \centering
    \includegraphics[scale=0.31]{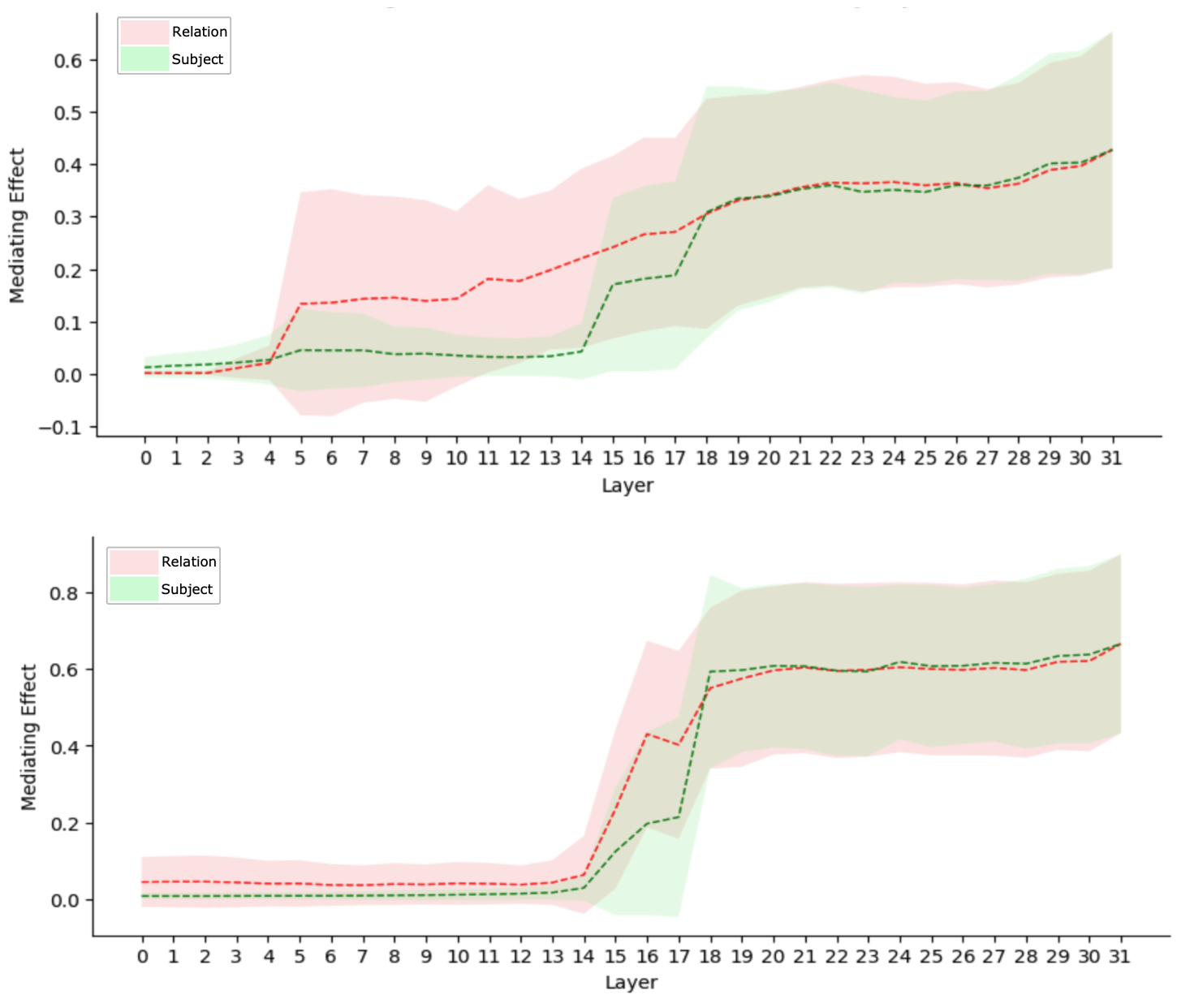}
    \caption{The mediating effects of subjects and relations of other two prompt templates. Top:\textit{"Tell me that the $<$relation$>$ of $<$subject$>$ is"}. Bottom:\textit{"If I want to know the $<$relation$>$ of $<$subject$>$, the answer is"}. The Three-Stage Observation is obvious.}
    \label{app_order}
    \vspace{-0mm}
\end{figure}

Figure \ref{app_order} shows the last position mediating effects of the subjects and relations of these two templates.
We find that the three-stage observation is not obvious as before.
Although the causal effect of relations still appear earlier than subjects, their gap is not so obvious.
We speculate that this may be caused by the position bias of LLMs. 
Some studies \cite{geva2023dissecting,xiao2023efficient} find that the allocation of attention is related to the order of token position. 
We leave to future research how attention allocation is related to the causal effects of subjects and relations.

\subsection{Verification by Early Decoding}
In this section, we use the early decoding technique to analyze what is encoded in the last token hidden states.
We map each hidden state $v^H_j,j\in[1,L]$ at layer $j$ as a distribution over the vocabulary by projecting it via the classification head $\phi$, followed by a softmax function.
\begin{equation}
        P_j = \mathit{Softmax}(\phi(v^H_j))
\end{equation}
Then we inspect the $k$ tokens with the highest probabilities in $P_j$.

Table \ref{lens_result1}, \ref{lens_result2}, \ref{lens_result3} show the early decoding result during fact recall processes in different relations, using Mistral-7b-instruct model. We select layers 2, 7, 16, and 27 for analysis, among which layer 2 belongs to the Initial stage, layer 7 and 16 belong to the Relational Emergence stage, and layer 27 belongs to the Conjoint influence stage.

We found that tokens obtained by decoding the hidden state of layer 2 are all meaningless tokens, which means that these hidden states do not encode useful information.
Some tokens obtained by decoding the hidden states of layer 7 and 16 exhibit the concept of relations, indicating that these hidden states solely encode relational representations.
Top-1 token obtained by decoding the hidden state of layer 27 is the correct object, indicating that these hidden states encode both subject and relation representations.
These early decoding results are consistent with our three-stage observation

\subsection{More Experimental Details}
\subsubsection{Data filtering}
Since our motivating observation is based on casual mediation analysis, which requires that the next output token of LLMs must be the correct object. 
Furthermore, in our designed hidden states transplantation experiment, it is essential that the object can be accurately predicted. 
This ensures that we can effectively assess whether the failure of object prediction after transplantation is due to incomplete knowledge of LLMs or if our hypothesis is wrong.

During our experiments, we collect the data from the factual set and commonsense set of this paper\cite{hernandez2023linearity} and filtered them to make sure the predicted next token is the object.
The filtered data details are show in Figure\ref{data_details}.
We apply this filtered dataset in the controllable fact recall experiments.

\begin{table}[h]
\caption{The initial quantity and filtered quantity of each relation type. The filtered objects can be predicted as the next token by the prompt \textit{"Given $<$subject$>$, the $<$relation$>$ of this one is"}}
\label{data_details}
\resizebox{0.5\textwidth}{!}{
\begin{tabular}{llll}
\hline
\multirow{2}{*}{Subject-Object} & \multirow{2}{*}{Initial quantity} & \multicolumn{2}{c}{Filtered quantity} \\
                                &                                   & Mistral-7b-insturct       & Llama-7b-chat      \\ \hline
Country-Capital                 & 253                               & 230              & 230                \\
Country-language                & 24                                & 24               & 24                 \\
Singer-Band                     & 21                                & 18               & 15                 \\
Landmark-Country                & 836                               & 637              & 647                \\
Country-Currency                & 30                                & 28               & 30                 \\
Person-Instrument               & 513                               & 280              & 365                \\
Pokemon-evolutions              & 44                                & 8                & 25                 \\
Product-Company                 & 522                               & 256              & 246                \\
Work-Location                   & 38                                & 26               & 27                 \\
Star-Constellation              & 362                               & 157              & 249                \\
Name-Religion                   & 31                                & 26               & 28                 \\
Company-Headquarters            & 674                               & 318              & 301                \\
Food-Country                    & 30                                & 22               & 24                 \\
Food-Color                      & 30                                & 15               & 19                 \\
Substance-Phase                 & 50                                & 43               & 36                 \\
Company-CEO                     & 298                               & 67               & 84                 \\
Superohero-True name            & 100                               & 42               & 50                 \\
Superhero-Enemy                 & 96                                & 24               & 22                 \\
Person-Language                 & 919                               & 802              & 810                \\
Athlete-Sport                   & 318                               & 239              & 250                \\
City-Country                    & 27                                & 27               & 24                 \\
Adj-Antonym                     & 100                               & 70               & 70                 \\ \hline
\end{tabular}
}
\end{table}
\subsubsection{Prompt Construction Details}
\paragraph{Hidden states transplantation}
In hidden states transplantation experiment, we need to construct reference prompts and source prompts, that contains different subjects and relations.
In each relation type, we first select 11 subjects, one of which is combined with the original relation to form a reference prompt, and the other 10 subjects are combined with other arbitrary relations to form source prompts.
Next, another subject is taken as the reference subject, while the other subjects serve as source objects.
This process is iterated to ensure the reference subjects and source subjects should have no intersection, which is important for us to determine whether the prediction of the source object is caused by the transplantation of the relational representation.
Table \ref{transplantation examples} shows some example reference and source prompts.

\paragraph{Controllable fact recall}
In the experiments of controllable fact recall, we need to combine subjects with other relations to construct various prompts for relation rewriting. 
Therefore, we collect a relation set with 100 relations as shown in Table \ref{setrandom}.
In the experiment, we select 10 subjects of the same relation type and combine them with the relation set to form 1000 inquiry prompts, and the accuracy is calculated average these 10 subjects. The inquiry prompt templates are shown in Table \ref{inquiry templates}.

\begin{table}[h]
\caption{The inquiry prompt templates in controllable fact recall experiments.}
\label{inquiry templates}
\begin{tabular}{l}
\hline
Can you tell me the $<$relation$>$ of $<$subject$>$? \\
What is the $<$relation$>$ of $<$subject$>$?         \\
Tell me the $<$relation$>$ of $<$subject$>$?         \\
I want to know the $<$relation$>$ of $<$subject$>$.  \\ \hline
\end{tabular}
\end{table}

\begin{table*}[b]
\caption{The early decoding of the prompt \textit{"Given France, the capital of this country is"}. All tokens are sorted from large to small according to probability.}
\label{lens_result1}
\resizebox{\textwidth}{!}{
\begin{tabular}{cl}
\hline
Layer & \multicolumn{1}{c}{Top-scoring Tokens}                                                                                                         \\ \hline
2     & kennis, bewerken, Namen, consid, weit, bekan, Hace, noten, Opcode, Autow, Erst, NOT, accomp, Opcode, Autow, accomp, sugg, pob \\ 
7     & NOT,  \underline{city}, definitely, not, neither, NO, miles, almost, unk, Ad, pres, particip, mie, wealth, \underline{cities}, closest, utt, \underline{home}, divided, wealth                         \\
16    & \underline{city}, \underline{cities}, NOT, cent, \underline{located}, not, \underline{London}, surely, None, also, definitely, typically, exactly, centers, \underline{towns}, \underline{City}, \underline{Capital}, always, nicht, obviously        \\
27    & \underline{Paris}, France, French, Berlin, Vers, rench, Madrid, Par, Jacques, Pierre, PAR, Washington, London, Seine, Orleans, city, Tokyo, Rome, capital, Wars               \\ \hline
\end{tabular}
}
\end{table*}

\begin{table*}[b]
\caption{The early decoding of the prompt \textit{"Given banana, the color of this food is"}.All tokens are sorted from large to small according to probability.}
\label{lens_result2}
\resizebox{\textwidth}{!}{
\begin{tabular}{cl}
\hline
Layer & \multicolumn{1}{c}{Top-scoring Tokens}                                                                                                          \\ \hline
2     & kennis, bewerken, Namen, accomp, Hace, Familie, INCLUDING, nt, CP, plaat, sugg, NOT, klik, membre, neces, efect, Schaus, comun, NOT, referenties         \\
7     & NOT, nt, dispon, definitely, \underline{purple}, uty, due, enu, probably, ED, \underline{colour}, NaN, pak, streak, everywhere, \underline{colours}, propriet, oct, gonna, going                       \\
16    & \underline{yellow}, \underline{color}, \underline{purple}, \underline{redd}, \underline{colors}, \underline{colour}, \underline{gray}, primarily, \underline{orange}, usually, typically, \underline{color}, \underline{Color}, dependent, \underline{brown}, \underline{green}, \underline{black}, \underline{colored}, \underline{colours} \\
27    & \underline{yellow}, brown, ban, Yellow, ri, green, orange, golden, mostly, usually, purple, Ban, mainly, Sei, Brown, typically, cream, tan, brow, redd, ban, generally           \\ \hline
\end{tabular}
}
\end{table*}

\begin{table*}[]
\caption{The early decoding of the prompt \textit{"Given Roberto Clemente, the sport that this athlete plays is"}.All tokens are sorted from large to small according to probability.}
\label{lens_result3}
\resizebox{\textwidth}{!}{
\begin{tabular}{cl}
\hline
Layer & \multicolumn{1}{c}{Top-scoring Tokens}                                                                                                                   \\ \hline
2     & CP, NOT, nt, consid, bewerken, accomp, lam, Z, pol, monument, weit, plaat, ebenfalls, noten, equal, dav, gouver, meant, pal, arqu                                  \\
7     & NOT, nt, nam, not, lan, diffusion, mitt, rl, influenced, prevention, probably, disp, both, Backend, definitely, neither, elf, ister, SET, n                                 \\
16    & nt, NOT, primarily, \underline{football}, rael, Unknown, \underline{basketball}, not, ohl, \underline{tennis}, None, itz, \underline{cricket}, Histor, unknown, elligence, definitely, becoming, none, namely               \\
27    & \underline{baseball}, Base, basketball, base, Major, soccer, professional, cricket, mostly, football, tennis, ML, hockey,Professional, American, golf,Puertoly,Sei,Brown \\ \hline
\end{tabular}
}
\end{table*}

\begin{table*}[]
\caption{The relation set for random combination.}
\label{setrandom}
\resizebox{\textwidth}{!}
{
\begin{tabular}{l}
\hline
"the size of", the weight of", "the height of", "the length of", "the width of", "the temperature of",\\ "the speed of", "the duration of", "the age of", "the volume of", "the density of",                                  
"the brightness of",\\"the intensity of", "the depth of", "the pressure of", "the sound of", "the texture of",                               
"the smell of", \\"the taste of", "the shape of", "the color of", "the pattern of", "the frequency of",                                      
"the resolution of",\\"the power of", "the energy of", "the efficiency of", "the accuracy of", "the precision of",                           
"the complexity of", \\"the simplicity of", "the clarity of", "the transparency of", "the opacity of", "the fragility of",                   
"the flexibility of",\\ "the rigidity of", "the brittleness of", "the hardness of", "the softness of","the future of","the acidity of"       \\
"the smoothness of", "the roughness of", "the elasticity of", "the conductivity of","the resistance of", "the magnetism of",               \\
"the gravity of", "the buoyancy of", "the transparency of", "the reflectivity of","the absorbency of", "the reactivity of",                \\
"the stability of", "the volatility of", "the solubility of", "the viscosity of","the alkalinity of", "the composition of",                \\
"the structure of", "the organization of", "the arrangement of","the formation of", "the development of", "the growth of",                 \\
"the evolution of", "the behavior of", "the function of","the purpose of", "the role of", "the significance of",                           \\
"the impact of", "the influence of", "the effect of", "the result of","the outcome of", "the consequence of",                              \\
"the importance of", "the relevance of", "the connection of", "the relationship of","the interaction of", "the correlation of",            \\
"the similarity of", "the difference of", "the contrast of", "the comparison of","the similarity of", "the variation of", \\
"the change of", "the improvement of", "the innovation of", "the advancement of","the discovery of", "the invention of", \\
"the application of", "the utilization of", "the adaptation of", "the transformation of"                                                                         \\ \hline
\end{tabular}
}
\end{table*}



\begin{table*}[]
\caption{Examples of source and references prompts in hidden states transplantation.}
\label{transplantation examples}
\resizebox{\textwidth}{!}{
\begin{tabular}{llll}
\hline
Relation    & Reference Prompts   & Source Prompts       & Reference/Source Targets     \\ \hline
Food-Color  & Given milk, the color of this food is                                                                & \begin{tabular}[c]{@{}l@{}}Given blueberry, the taste of this food is\\ Given cucumber, the texture of this food is\\  Given apple, the shape of this food is\\ Given chocoloate, the weight of this food is\\ Given watermelon, the shape of this food is\\ Given banana, the height of this food is\\ ...\end{tabular}                                                                                                               & \begin{tabular}[c]{@{}l@{}}White/Blue\\ White/Green\\ White/Red\\ White/Brown\\ White/Green\\ White/Yellow\\ ...\end{tabular}                                           \\ \hline
Singer-Band & \begin{tabular}[c]{@{}l@{}}Given Brian Johnson, the band name \\ of this lead singer is\end{tabular} & \begin{tabular}[c]{@{}l@{}}Given Ozzy Osbourne, the nationality of this lead singer is\\ Given Robert Plant, the hobby of this lead singer is\\ Given Liam Gallagher, the language of this lead singer is\\ Given Eddie Vedder, the nationality of this lead singer is\\ Given Billie Joe Armstrong, the age of this lead singer is\\ Given Chris Martin, the homeland of this lead singer is\\ ...\end{tabular} & \begin{tabular}[c]{@{}l@{}}"AC/DC"/Black Sabbath\\  "AC/DC"/Led Zeppelin\\ "AC/DC"/Oasis\\  "AC/DC"/Pearl Jam\\ "AC/DC"/Green Day\\ "AC/DC"/Coldplay\\ ...\end{tabular} \\ \hline
\end{tabular}
}
\end{table*}

\begin{table*}[t]
\vspace{0mm}
\caption{We use the relation \textit{the capital of} to control the generation about \textit{France} and \textit{Italy}}
\label{generation result1}
\resizebox{\textwidth}{!}{
\begin{tabular}{|l|l|l|}
\hline
Prompt           & Original Output         & Relational Rewritten Output       \\ \hline
\begin{tabular}[c]{@{}l@{}}Please introduce some \\ information about France.\end{tabular} 
& \begin{tabular}[c]{@{}l@{}}France is a country located in \\Western Europe and has a population \\ of around 67 million people. Its \\official language is French, and its \\capital and largest city is  Paris. France \\is known for its rich history,  art, \\ culture, and cuisine. The country \\ has a diverse landscape……\end{tabular}                        
& \begin{tabular}[c]{@{}l@{}}Paris, also known as the City of Light, \\is the capital and largest city of France. \\It is located in the north-central region \\ of the country along the Seine River and \\is one of the most visited cities in the \\world. Paris is known for its stunning \\architecture...\end{tabular}                                                                          \\ \hline
\begin{tabular}[c]{@{}l@{}}Give me some information \\ about Italy.\end{tabular}           & \begin{tabular}[c]{@{}l@{}}Italy is a country located in\\  southwestern Europe. It is the seventh\\-largest country in the European\\ Union and the thirty-sixth largest \\country in the world.  Its population is \\approximately 60 million. The capital \\and largest city is Rome, with \\other major cities including Milan, \\Naples, Florence, and Turin. Italy \\is known for its diverse culture,\\ history, art......\end{tabular} & \begin{tabular}[c]{@{}l@{}}Rome is the capital of Italy and \\ one of the most ancient cities in \\ the world. It is located in the central-\\ western part of the Italian peninsula\\, on  the Tiber River, and has a \\ population of over 2.8 million people.\\Rome is known for its rich history and \\ cultural heritage, and is home to many \\famous landmarks and attractions\end{tabular} \\ \hline
\end{tabular}
}
\end{table*}

\begin{table*}[]
\caption{We use the relation \textit{the currency of} to control the generation about \textit{Korea} and \textit{Japan}}
\label{generation result2}
\resizebox{\textwidth}{!}{
\begin{tabular}{|l|l|l|}
\hline
Prompt           & Original Output         & Relational Rewritten Output       \\ \hline
\begin{tabular}[c]{@{}l@{}}Tell me some information \\ about Korea.\end{tabular} 
& \begin{tabular}[c]{@{}l@{}}Korea, officially the Republic of \\Korea (ROK), is a country in East \\ Asia.It is located on the Korean \\Peninsula, bordered by North Korea \\ to the north, China to the northeast, \\and the Sea of Japan to the east. The \\capital and largest city of Korea is \\ Seoul, located in the south-central\\ part of the peninsula. Korea has \\a population of over 51 million \\people and is the 21st most\\ populous country...\end{tabular}                        
& \begin{tabular}[c]{@{}l@{}}the Korean won.
The Korean \\won (KRW)  is the official currency \\of South  Korea. It is also used in \\North Korea, although the North \\Korean won is not recognized \\outside of North Korea. The Korean \\won   is divided into 10,000 won \\(KRW 10,000), 5,000 won \\(KRW 5,000), 2,000 won \\(KRW 2,000)...\end{tabular}                                                                          \\ \hline
\begin{tabular}[c]{@{}l@{}}Tell me some information \\ about Japan.\end{tabular}           & \begin{tabular}[c]{@{}l@{}}Japan, officially the Empire of Japan, \\is a country located in East Asia.\\ It is made up of four main islands: \\Hokkaido, Honshu, Shikoku, and \\Kyushu, and several smaller islands. \\The capital and largest city is Tokyo, \\while other major cities include \\Yokohama, Osaka, and Nagoya. \\Japan has a population of over \\126 million people and its official \\ language is Japanese. Japan has \\a rich history and culture, with \\influences from Buddhism...\end{tabular} & \begin{tabular}[c]{@{}l@{}}Yen. 1. What is the name of the\\ currency used in Japan? The \\currency used in Japan is the \\Japanese yen, often abbreviated as\\ JPY. 2. What is the symbol for\\ the Japanese yen? The symbol for \\the Japanese yen is \textyen. 3. What is the \\value of 1 Japanese yen? As of \\August 2021, 1 Japanese yen is \\equivalent to approximately \\0.0092 US dollars...
\end{tabular} \\ \hline
\end{tabular}
}
\end{table*}

\begin{table*}[]
\caption{We use the relation \textit{the taste of} to control the generation about \textit{milk} and \textit{cucumber}}
\label{generation result3}
\resizebox{\textwidth}{!}{
\begin{tabular}{|l|l|l|}
\hline
Prompt           & Original Output         & Relational Rewritten Output       \\ \hline
\begin{tabular}[c]{@{}l@{}}Tell me what you know \\ about milk.\end{tabular} 
& \begin{tabular}[c]{@{}l@{}}Milk is a nutrient-rich liquid\\ produced by the mammary glands \\of female mammals, primarily cows,\\ goats, and sheep. It is a natural\\ source of essential vitamins and \\minerals, including calcium, \\vitamin D, vitamin A, vitamin E, \\potassium, and vitamin B12....\end{tabular}                        
& \begin{tabular}[c]{@{}l@{}}sour, sweet, or bitter.\\
A: Milk can have different tastes \\depending on its origin and\\ processing. It can be sour due to \\the presence of lactic acid, sweet \\due to the presence of natural \\sugars, or bitter due to the \\presence of certain compounds...\end{tabular}                                                                          \\ \hline
\begin{tabular}[c]{@{}l@{}}Tell me what you know \\about cucumber.\end{tabular}           & \begin{tabular}[c]{@{}l@{}}Cucumber is a type of edible squash \\that belongs to the Cucurbitaceae \\family. It is a popular vegetable in \\many cuisines around the world,\\ known for its refreshing taste and \\mild crunch. Cucumbers are grown  \\in warm, humid climates and are \\harvested when they are fully ripe,\\ with the skin turning yellow or\\ green. They are low in calories \\and high in water content, making\\ them a healthy choice for people ...\end{tabular} & \begin{tabular}[c]{@{}l@{}}sweet and mildly bitter.\\
Cucumbers are a type of fruit \\that are known for their sweet \\and mildly bitter taste. They are \\often used in salads, sandwiches,\\ and as a snack. Cucumbers \\are also used in pickling and\\ preserving, and are a popular \\ingredient in many cuisines around \\the world. They are a good \\source of vitamins and minerals...
\end{tabular} \\ \hline
\end{tabular}
}
\end{table*}

\begin{table*}[]
\caption{We use the relation \textit{the CEO of} to control the generation about \textit{Microsoft} and \textit{Google}}
\label{generation result4}
\resizebox{\textwidth}{!}{
\begin{tabular}{|l|l|l|}
\hline
Prompt           & Original Output         & Relational Rewritten Output       \\ \hline
\begin{tabular}[c]{@{}l@{}}Share some facts about \\Microsoft with me.\end{tabular} 
& \begin{tabular}[c]{@{}l@{}}1. Microsoft was founded on \\April 4, 1975, by Bill Gates and \\ Paul Allen. 2. The company's \\ name is derived from the words \\"microcomputer" and "software." 3.\\ Microsoft's first product was the \\ Altair 8800, a microcomputer \\that was released in 1975.\\ 4. Microsoft's Windows operating\\ system was first released in 1985 \\and has since become the most\\ widely used operating system in \\the world. 5. Microsoft's Office....\end{tabular}                        
& \begin{tabular}[c]{@{}l@{}}Satya Nadella.
Satya Nadella is the \\ CEO of Microsoft. He was born \\ on September 28, 1968, in \\ Hyderabad, India. Nadella \\ received a Bachelor of Science\\  in Computer Science from the \\University  of Washington in 1990 \\ and a Master of Business \\Administration  from the Stanford \\Graduate  School of Business \\in 1997. 
Nadella joined Microsoft\\ in 1992 and has held several \\leadership  positions within the ...\end{tabular}                                                                          \\ \hline
\begin{tabular}[c]{@{}l@{}}Can you give me some \\details about Google?\end{tabular}           & \begin{tabular}[c]{@{}l@{}}
Google is a multinational technology \\company that specializes in Internet-\\related services and products. It was \\founded in 1998 by Larry Page and \\Sergey Brin, and is headquartered \\in Mountain View, California. \\Google's flagship product is its \\search engine, which is used by \\millions of people around the world \\to find information on the Internet. \\In addition to its search engine, \\Google also offers a wide range \\of other products and services...\end{tabular} & \begin{tabular}[c]{@{}l@{}}Sundar Pichai? Sundar Pichai is the \\CEO of Google, one of the world's \\largest technology companies.\\ He was born in India in 1971 and \\ moved to the United States with\\ his family when he was 13\\ years old. He studied computer \\science at Stanford University \\and later worked at Google as a \\software engineer before becoming \\the CEO in 2015. Pichai is known \\for his leadership in the tech \\industry and his commitment to....
\end{tabular} \\ \hline
\end{tabular}
}
\end{table*}

\end{document}